\documentclass[11pt,a4paper]{article}
\usepackage[hyperref]{acl2018}
\usepackage{graphicx}
\usepackage{amsmath}
\usepackage{paralist}
\usepackage{subcaption}

\aclfinalcopy 

\title{Judging a Book by its Description : {\it Analyzing Gender Stereotypes in the Man Bookers Prize Winning Fiction}}

\author{Nishtha Madaan\textsuperscript{1}, Sameep Mehta\textsuperscript{1}, Shravika Mittal \textsuperscript{2}, Ashima Suvarna\textsuperscript{3}\\ 
\textsuperscript{1}IBM Research AI-India\\
\textsuperscript{2}IIIT-Delhi, 
\textsuperscript{3}DTU-Delhi\\
  Contact: \{nishthamadaan, sameepmehta\}@in.ibm.com\\ 
  }
\date{}

\begin{document}
\maketitle

\begin{abstract}
The presence of gender stereotypes in many aspects of society is a well-known phenomenon. In this paper, we focus on studying and quantifying such stereotypes and bias in the Man Bookers Prize winning fiction. We consider 275 books shortlisted for Man Bookers Prize between 1969 and 2017. The gender bias is analyzed by semantic modeling of book descriptions on Goodreads. This reveals the pervasiveness of gender bias and stereotype in the books on different features like occupation, introductions and actions associated to the characters in the book.
\end{abstract}

\begin{figure*}[h]
 \includegraphics[width=1.0\linewidth]{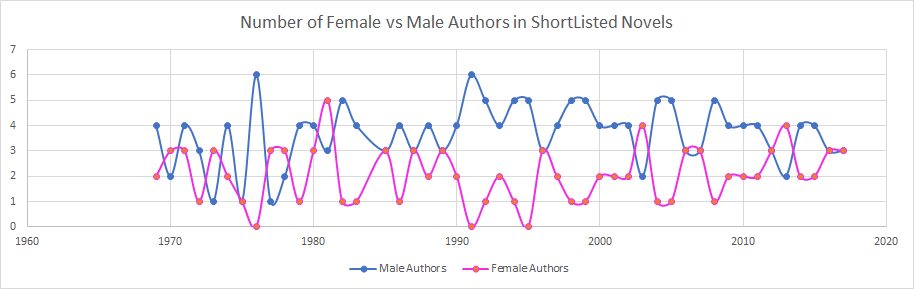}
  \caption{Male and Female Authors in Fiction. Count of Female Authors is shown in pink and Count of Male Authors is shown in blue.}
  \label{fig:authors}
\end{figure*}

\section{Introduction}

Gender, racial and ethnic stereotypes in many aspects of society is an undesirable yet pervasive phenomenon. In this work, we analyze and quantify gender-based stereotypes in description (summary) of fiction books shortlisted for Man Bookers Prize during the period 1969 to 2017. We show how gender bias and stereotyping is present in these books. The trends are improving but still far from optimal. We also look at gender of the authors which point to a gender imbalance problem shown in figure \ref{fig:authors}. However, in this work we focus on gender of characters more than gender of authors.

The motivation for considering books has been three fold: 
\begin{asparaenum}[a)]
    \item The data is very diverse in nature. Hence finding how gender stereotypes exist in this data becomes an interesting study.
    \item The data-set is large. We analyze 275 books which cover all the books shortlisted for Man Bookers prize since 1969.  So it becomes a good first step to develop computational tools to analyze the existence of stereotypes over a period of time.
    \item These books are a reflection of society. It is a good first step to look for such gender bias in this data so that necessary steps can be taken to remove these biases.
\end{asparaenum}

While many regular book readers would have had similar hunches, to best of our knowledge we are the first ones to use Text Analytics, NLP and Graph based algorithms to study this computationally.

We focus on following tasks to study gender bias in Man Bookers Winning Fiction. 
\begin{asparaenum}[I)]
    \item \textbf{Occupations and Gender Stereotypes}- How are males portrayed in their jobs versus females? How are these levels different? How does it correlate to gender bias and stereotype? As shown in previous studies done on Hollywood and Bollywood story plots and scripts. Gender stereotyping with respect to occupations is one of the most pervasive biases that cuts across countries and age groups. This is evidenced by our previous work analyzing Bollywood movie story-lines \cite{madaan2018analyze}.
    \item \textbf{Appearance} - How are males and females described on the basis of their appearance? How do the descriptions differ in both of them? How does that indicate gender stereotyping?
    \item \textbf{Mentions} - How many males and females are mentioned in the fiction? 
    \item \textbf{Descriptions} - How do the descriptions of a male and a female differ in the books?
\end{asparaenum}


Detection of such bias is only the first step. We are also developing various algorithms to debias such text. We have developed a focused de-biaser with respect to gender stereotyping in occupations \cite{madaan2018generating}. The occupation bias has also been recently noted in machine translation systems \cite{caliskan2016semantics}.

In parallel, we are also working on generalized reasoning based algorithm DeCogTeller to debias complete text. Early results are positive and encouraging!


\begin{figure*}
\begin{subfigure}{0.5\textwidth}
 \includegraphics[width=\linewidth]{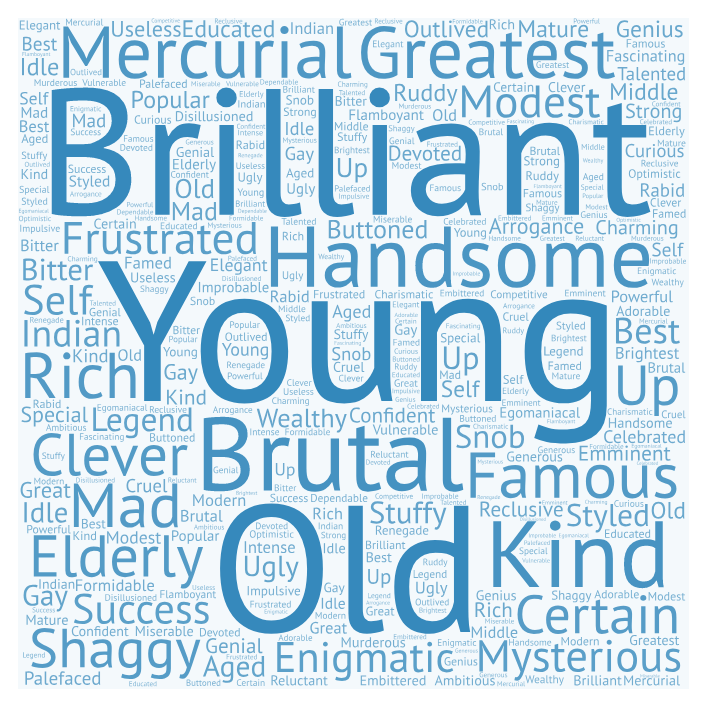}
\end{subfigure}\hspace*{\fill}
\begin{subfigure}{0.5\textwidth}
 \includegraphics[width=\linewidth]{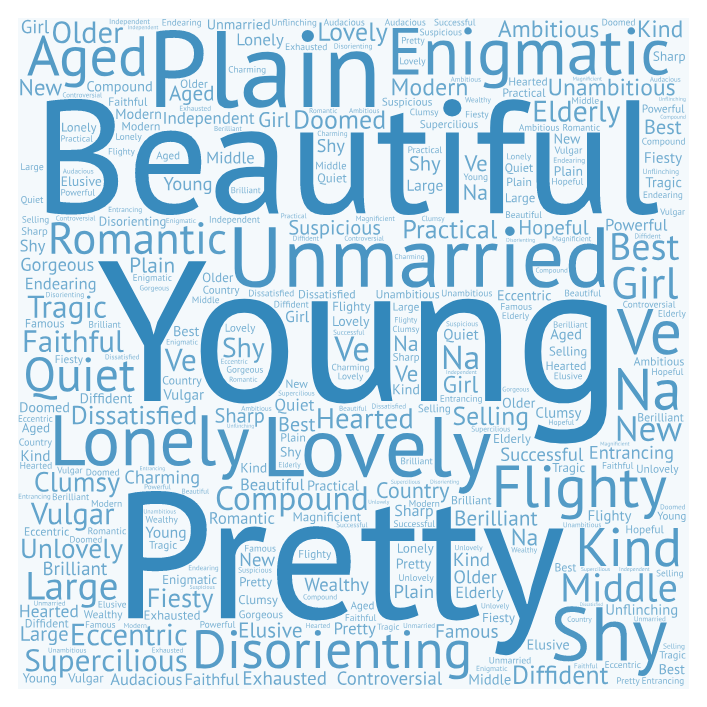}
\end{subfigure}\hspace*{\fill}
    \caption{Adjectives used with males and females}
    \label{fig:adj}
\end{figure*}
\begin{figure*}
\begin{subfigure}{0.5\textwidth}
 \includegraphics[width=\linewidth]{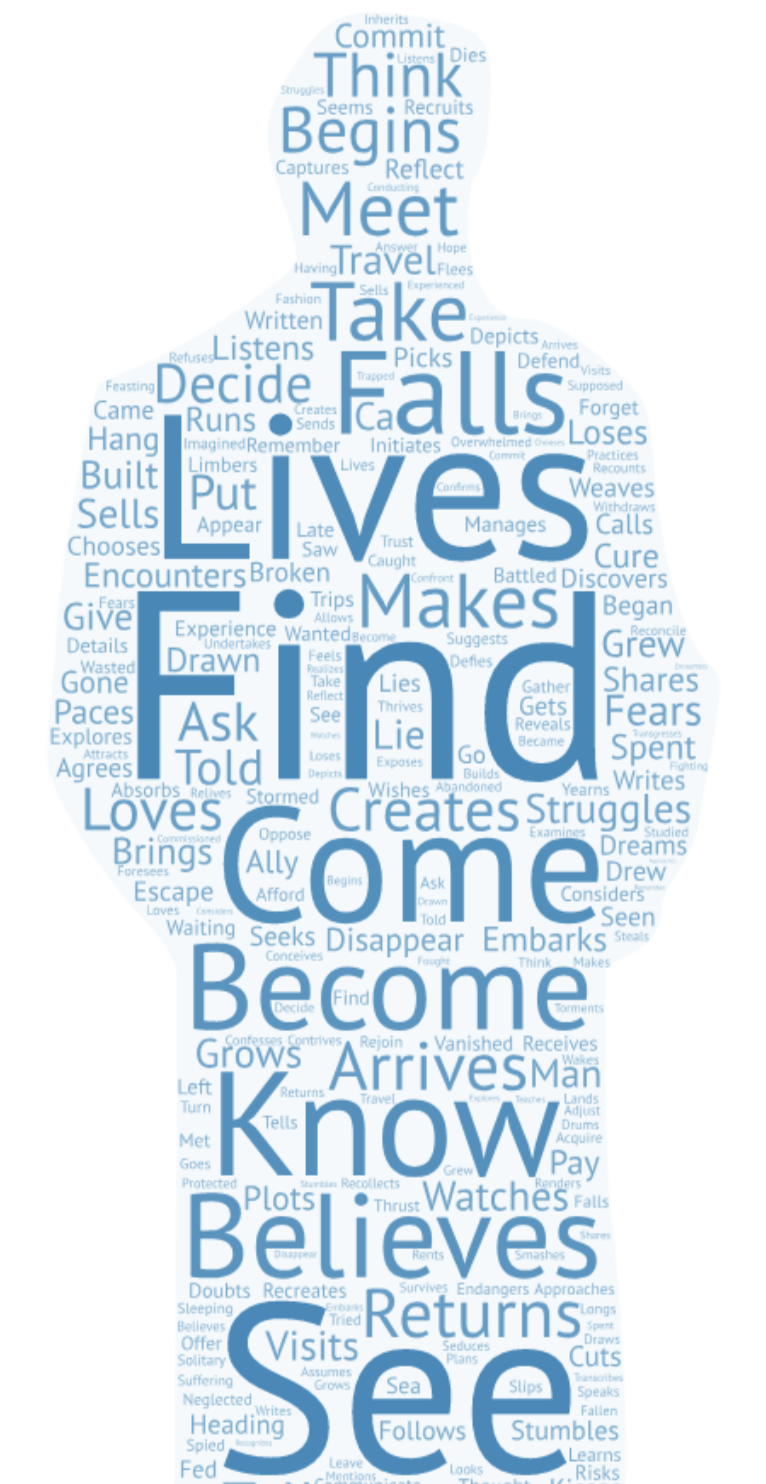}
\end{subfigure}\hspace*{\fill}
\begin{subfigure}{0.5\textwidth}
 \includegraphics[width=\linewidth]{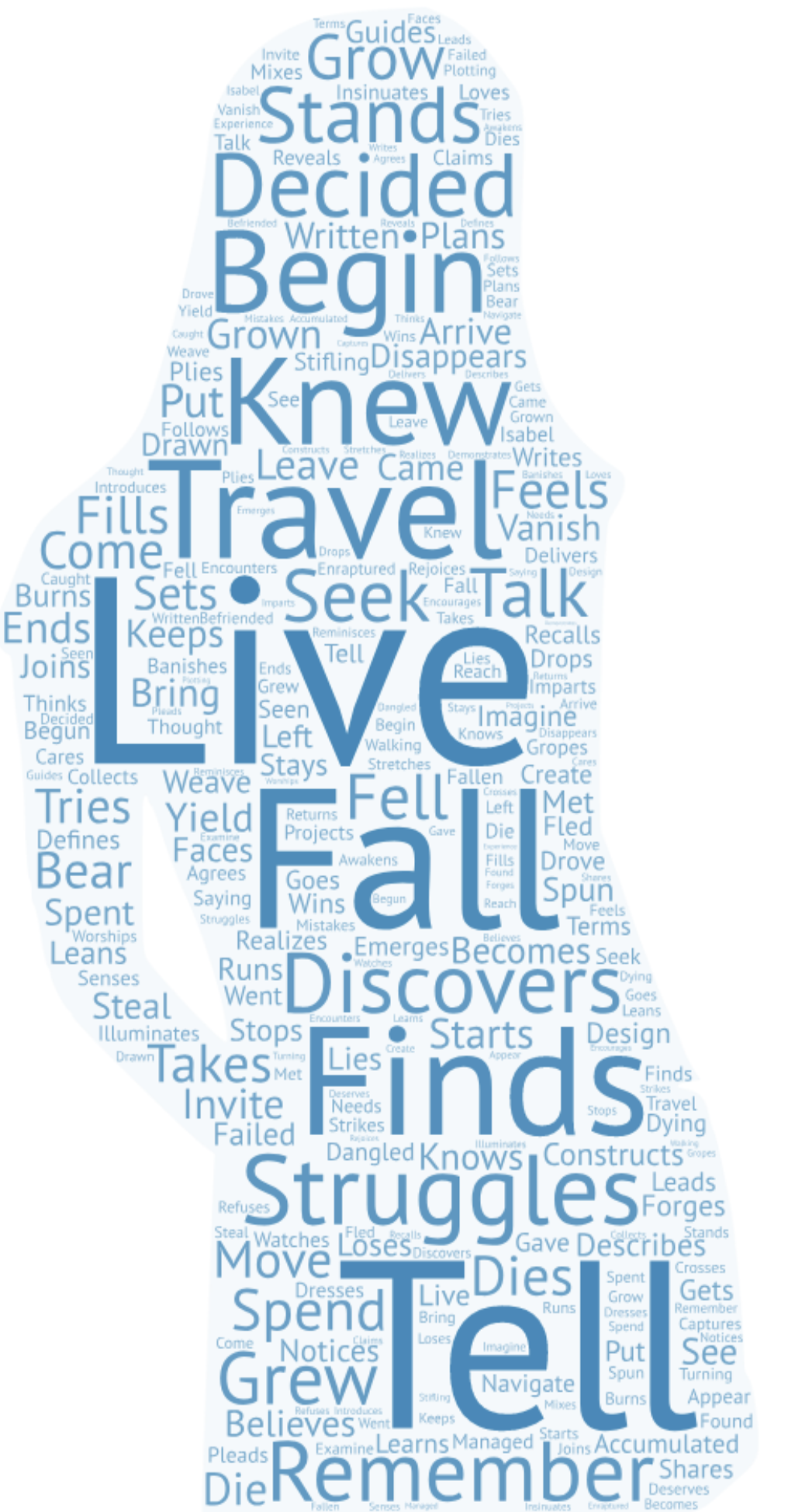}
\end{subfigure}\hspace*{\fill}
    \caption{Verbs used with males and females}
    \label{fig:verbs}
\end{figure*}
\begin{figure*}
\begin{subfigure}{0.5\textwidth}
 \includegraphics[width=\linewidth]{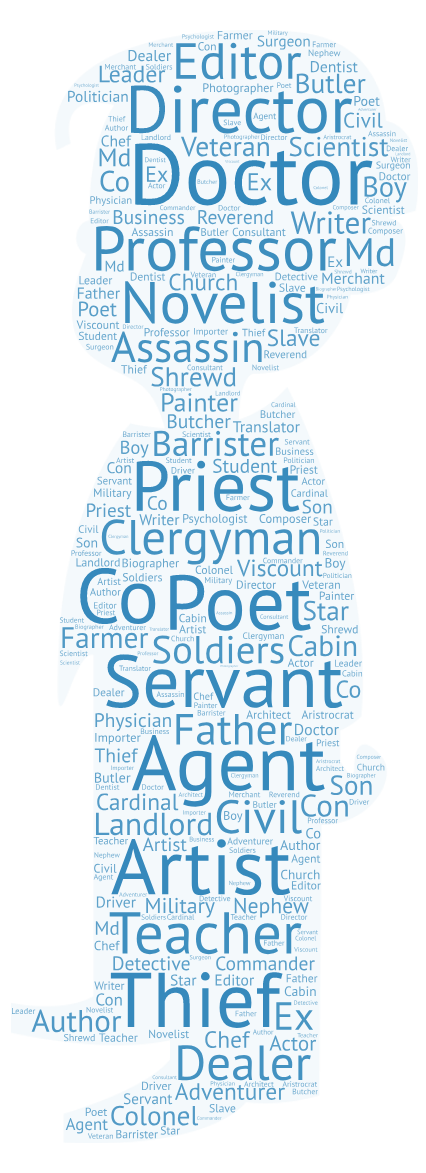}
\end{subfigure}\hspace*{\fill}
\begin{subfigure}{0.5\textwidth}
 \includegraphics[width=\linewidth]{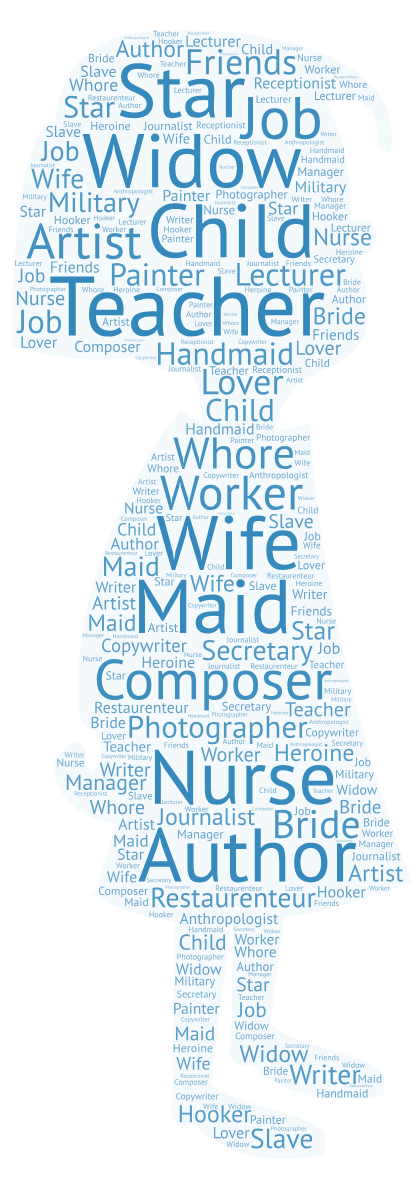}
\end{subfigure}\hspace*{\fill}
    \caption{Occupations of males and females}
    \label{fig:occupations}
\end{figure*}

\section{Past Work}
\textbf{Analysis of gender bias in machine learning} in recent years has not only revealed the prevalence of such biases but also motivated much of the recent interest and work in de-biasing of ML models. \cite{zhao2017men} have pointed to the presence of gender bias in structured prediction from images. \cite{fast2016shirtless, madaan2018analyze} notice these biases in movies while \cite{gooden2001gender, millar2008selective} notice the same in children books and music lyrics.

While there are recent works where gender bias has been studied in different walks of life \cite{soklaridis2017gender},\cite{ macnell2015s}, \cite{carnes2015effect}, \cite{terrell2017gender}, the analysis majorly involves information retrieval tasks involving a wide variety of prior work in this area. \cite{fast2016shirtless} have worked on gender stereotypes in English fiction particularly on the Online Fiction Writing Community. The work deals primarily with the analysis of how males and females behave and are described in this online fiction. Furthermore, this work also presents that males are over-represented and finds that traditional gender stereotypes are common throughout every genre in the online fiction data used for analysis. \\ Apart from this, various works where Hollywood movies have been analyzed for having such gender bias present in them \cite{blog}. Similar analysis has been done on children books \cite{gooden2001gender} and music lyrics \cite{millar2008selective} which found that men are portrayed as strong and violent, and on the other hand, women are associated with home and are considered to be gentle and less active compared to men. These studies have been very useful to uncover the trend but the derivation of these analyses has been done on very small data sets. In some works, gender drives the decision for being hired in corporate organizations \cite{dobbin2012corporate}. Not just hiring, it has been shown that human resource professionals' decisions on whether an employee should get a raise have also been driven by gender stereotypes by putting down female claims of raise requests. While, when it comes to consideration of opinion, views of females are weighted less as compared to those of men \cite{otterbacher2015linguistic}. On social media and dating sites, women are judged by their appearance while men are judged mostly by how they behave \cite{rose2012face,otterbacher2015linguistic,fiore2008assessing}. When considering occupation, females are often designated lower level roles as compared to their male counterparts in image search results of occupations \cite{kay2015unequal}. 

\subsection{Debiasing Algorithms}

\textbf{De-biasing the training algorithm} as a way to remove the biases focuses on training paradigms that would result in fair predictions by an ML model. In the Bayesian network setting, Kushner et al. have proposed a latent-variable based approach to ensure counter-factual fairness in ML predictions.  Another interesting technique (\cite{beutel2013copycatch} and \cite{zhang2016adversarial}) is to train a primary classifier while simultaneously trying to "deceive" an adversarial classifier that tries to predict gender from the predictions of the primary classifier.

\textbf{De-biasing the model after training} as a way to remove bias focuses on "fixing" the model after training is complete. \cite{bolukbasi2016man} in their famous work on gender bias in word embeddings take this approach to "fix" the embeddings after training.

\textbf{De-biasing the data at the source} fixes the data set before it is consumed for training. This is the approach we take in this paper by trying to de-bias the data or suggesting the possibility of de-biasing the data to a human-in-the-loop. A related task is to modify or paraphrase text data to obfuscate gender as in \cite{reddy2016obfuscating} Another closely related work is to change the style of the text to different levels of formality as in \cite{rao2018dear}.

{\it Please note that most of these approaches are proposed for numerical data. Detecting and De-biasing text is an upcoming area with very less work till now.}

\section{Data and Experimental Study}
\subsection{Data}
 \begin{figure*}[h]
 \includegraphics[width=1.0\linewidth]{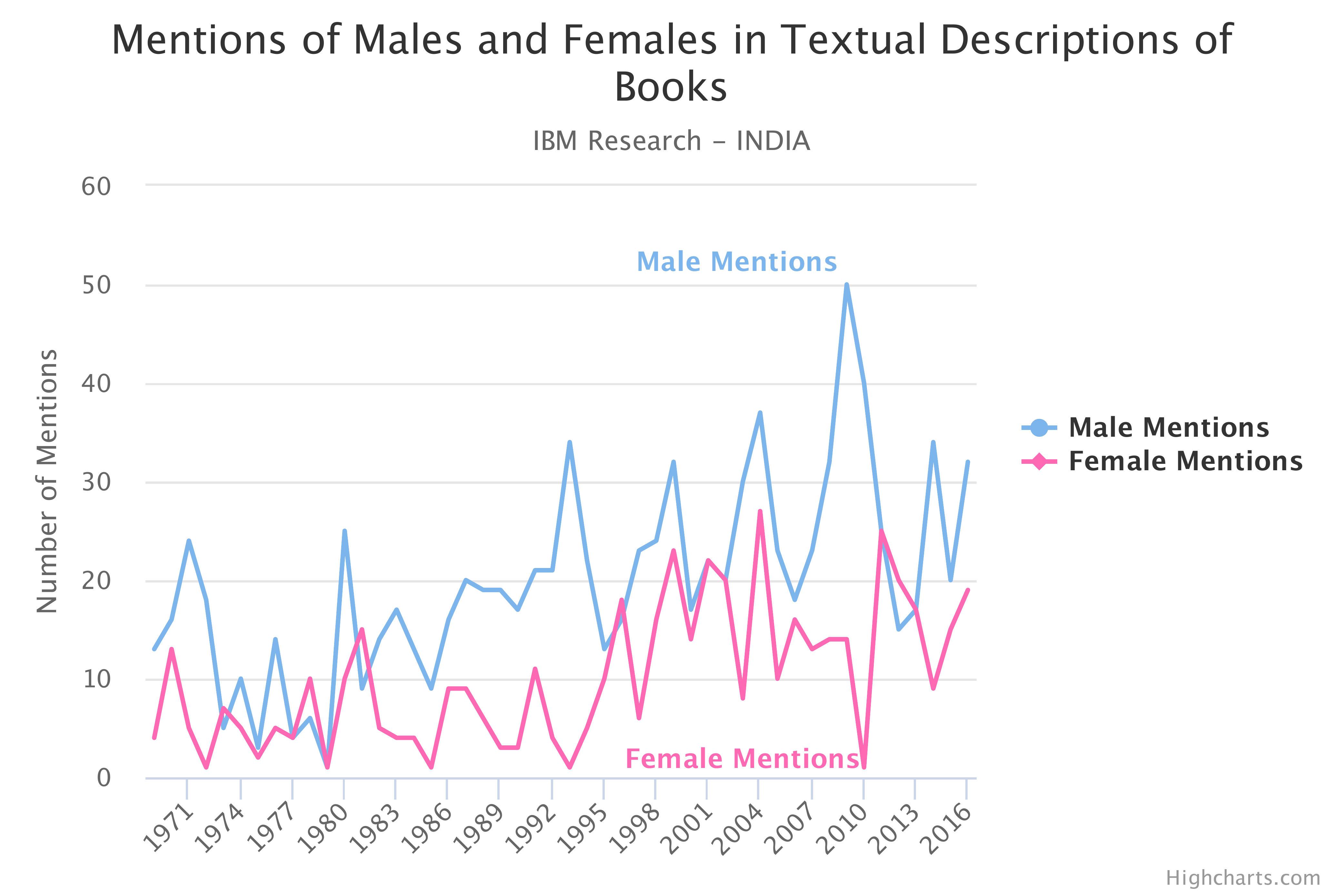}
  \caption{Total Character Mentions showing mentions of male and female characters. Female mentions are presented in pink and Male mentions in blue}
  \label{fig:mentions}
\end{figure*}
 The data-set contains 275 books for 1969-2017 time period. For each year we consider short listed books. The data-set consist of textual descriptions of books shortlisted for Man Bookers Prize from Goodreads. This text data is analysed using text analytics algorithms as explained in next section.

\subsection{Task and Approach}
In this section, we discuss the tasks we perform on the books data extracted from Goodreads. Further, we define the approach we adopt to perform individual tasks and then study the inferences. We define different tasks corresponding to our analysis.


To make books analysis ready, we used OpenIE \cite{fader2011identifying} for performing co-reference resolution on books text.  Co-reference task involves finding all expressions in text which maps to same entity. For example, consider a small snippet -- {\tt John went to market. He bought fruits.} In these sentence co-reference will map He to John. The co-referenced textual description is used for following analyses.

 \begin{asparaenum}[1)]
    \item \textbf{Character Mentions in the Book Descriptions} - We extracted mentions of male and female characters in the books description. The motivation to find mentions is how many times males have been referred to in the book versus how many times females have been referred to in the book. This helps us identify if the female has an important role in the book or not. In Figure \ref{fig:mentions} it is observed that, a male is mentioned around 30 times in a book while a female is mentioned only around 15 times. {\it Moreover, there is a consistency of this ratio from 1969 to 2017(for almost 50 years)!}
 
    \item \textbf{Character Appearance in Books Data} - We analyzed how male characters and female characters have been addressed. This involves extracting adjectives associated with male characters and female characters. To extract adjectives linked to a particular character, we use IBM Watson Natural Language Understanding API \cite{ibm}. In Fig \ref{fig:adj} we present the adjectives associated with males and females.  When we look at adjectives, males are often represented as rich and wealthy while females are represented as beautiful and attractive in books description. 
    
    \item \textbf{Character Descriptions in Books Data} - We analyze how the male characters and the female characters have been introduced in the textual description. This involves extracting verbs associated with the male and female characters. To extract verbs linked to a particular character, we use Stanford POS tagger \cite{de2006generating}. In Fig \ref{fig:verbs} we present the verbs associated with males and females. 
     When we look at verbs, males are often represented as powerful while females are represented as fearful.
 
    \item \textbf{Occupation as a stereotype} - We perform a study on how occupations of males and females are represented. To perform this analysis, we collated an occupation list from multiple sources over the web comprising of ~350 occupations. We then extracted an associated "noun" tag attached with character member of the story using Stanford Dependency Parser \cite{de2006generating} which is later matched to the available occupation list. In this way, we extract occupations for each character. We group these occupations for male and female characters for all the collated books data. Figure \ref{fig:occupations} shows the occupation distribution of males and females. From the figure it is clearly evident that, males are given higher level occupations than females. Our analysis shows that when it comes to occupation like "teacher" or "whore", females are high in number. But for "professor" and "doctor" the story is totally opposite.Detailed occupations are shown in table \ref{tab:tab1}
    
\begin{table*}
\centering
\begin{tabular} {|p{0.5\linewidth}|p{0.5\linewidth}|}
\hline
 \textbf{Top Occupations in Males} & \textbf{Top Occupations in Females} \\ \hline
Doctor/Physician/Surgeon/Psychologist & Teacher/Lecturer\\ \hline
Professor/Scientist & Nurse\\ \hline
Business/Director & Whore/Hooker\\ \hline
Church Agent/ Clergymen & Child wife/ Child Bride\\ \hline
Poet & Maid\\ \hline
Thief & Secretary\\ \hline
\end{tabular}
\caption{Occupations in Male and Female Characters in Books}
\label{tab:tab1}
\end{table*}


\section{Wind of Change}
Our system discovered at least 6-7 books in last four years where females play central role in textual description of the story. Few notable examples being - \tt{Do Not Say We Have Nothing} written by- {\it Madeleine Thien}, \tt{How to be Both} written by- {\it Ali Smith}, \tt{We Are All Completely Beside Ourselves} written by- {\it Karen Joy Fowler}, \tt{Eileen} written by- {\it Ottessa Moshfegh}, \tt{We Need New Names} written by- {\it NoViolet Bulawayo}, \tt{A Spool Of Blue Thread} written by-{\it Anne Taylor}, \tt{The Lowland} written by-{\it Jhumpa Lahiri}. We also note that over time such biases are decreasing - still far away from being neutral but the trend is encouraging. {\it Incidently all these books are written by female authors!}.

\end{asparaenum}

\section{Bias Removal Tool- DeCogTeller}
The system enables the user to enter some biased text and generate unbiased version of that text snippet. For this task, we take a news articles data set and train word embedding using Google \emph{word2vec} \cite{mikolov2013efficient}. This data acts as a \emph{fact data} which is used later to check for gender specificity of a particular action as per the facts. Apart from interchanging the actions, we have developed a specialized module to handle occupations. Very often, gender bias shows in assigned occupation \{ (Male, Doctor), (Female, Nurse)\} or \{ (Male, Boss), (Female, Assistant)\}. 


We give a holistic view of our system which is described in a detailed manner as follows-
\begin{asparaenum}[I)]
\item \textbf{Data Pre-processing} - We perform data pre-processing of the words in the fact data. (a) We look search Wordnet \cite{miller1995wordnet} to find whether the word in the fact data is present, and remove the word if not found. (b) We perform word stemming using the Stanford stemmer.

\item \textbf{Generating word vectors} - We train Google word2vec on the pre-processed data, and generate word embedding.

\item \textbf{Extraction of analogical pairs} - The next task is to find analogical pairs from fact data which are analogous to the $(man,woman)$ pair. E.g., if we take an analogical word pair $(x,y)$ and we associate a vector $P(x,y)$ to the pair, then, representing man as {\it m} and woman as {\it w}, the task is to find
\begin{equation}
P(\vec{\textbf{x}},\vec{\textbf{y}}) = (\vec{\textbf{m}} - \vec{\textbf{w}}) - (\vec{\textbf{x}} - \vec{\textbf{y}})
\label{eqn:manwoman}
\end{equation}

In Equation~\ref{eqn:manwoman}, if we replace man and woman vectors by \emph{he} ($\vec{\textbf{h}}$) and \emph{she} ($\vec{\textbf{s}}$) respectively, the above equation becomes
\begin{equation}
P(\vec{\textbf{x}},\vec{\textbf{y}}) = (\vec{\textbf{he}} - \vec{\textbf{she}}) - (\vec{\textbf{x}} - \vec{\textbf{y}})
\label{eqn:heshe}
\end{equation}

The intent is to capture word pairs such as doctor or nurse where in most of the data, doctor is close to he and nurse is closer to she. Therefore for $(x,y) = (doctor,nurse)$, we get
$$P(\vec{\textbf{doctor}},\vec{\textbf{nurse}}) \leftarrow (\vec{\textbf{he}} - \vec{\textbf{she}}) - (\vec{\textbf{doctor}} - \vec{\textbf{nurse}})$$

Another example of $(x,y)$ found in our data is $(king,queen)$.
We generate all such $(x,y)$ pairs and store them in our knowledge base. To have refined pairs, we used a scoring mechanism to filter important pairs. If
\[
\lVert  \mathbf P(\vec{\textbf{x}},\vec{\textbf{y}}) \rVert \leq \tau
\]
where $\tau$ is the threshold parameter, then add the word pair to knowledge base otherwise ignore. Equivalently, after normalizing $(\vec{\textbf{he}}-\vec{\textbf{she}})$ and $(\vec{\textbf{x}}-\vec{\textbf{y}})$, we calculated cosine distance as $cosine(\vec{\textbf{he}} - \vec{\textbf{she}}, \vec{\textbf{x}} - \vec{\textbf{y}})$, which is algebraically equivalent to the above inequality. 

In our system, we extract plausible analogical word pairs by selecting candidates (the $\vec{\textbf{x}}$ and $\vec{\textbf{y}}$ described in Equation~\ref{eqn:heshe}) for each character appearing in the sentence, jointly using IBM's and UIUC's semantic role labeler \cite{ibm}\cite{PunyakanokRoYi08}, and picking the objects associated with that character via some labeled role.

\item \textbf{Classifying word pairs} - 

 \textbf{Introducing word pair interchangeability}- A pair of words are interchangeable for gender, if their roles, actions or relationships can be exchanged without breaking gender-related practical plausibility. For instance, in the pair \emph{(doctor, nurse)}, being a {\it doctor} and a {\it nurse} are gender-neutral roles, so the word pair can be interchanged. Contrarily, in \emph{(king, queen)}, such interchange is non-plausible (male queens and female kings are non-plausible).

 \textbf{Performing interchange}- In order to perform word pair interchange, we determine which pairs extracted in the above step correspond to gender neutral and which ones correspond to gender specific. To do this, we first extract the words from knowledge base extracted from test data and find how close they are to different genders. We find the cosine distance of the words in the word pair with $\vec{\textbf{he}}$ and $\vec{\textbf{she}}$ respectively, and if any word is close enough within a threshold to any of $\vec{\textbf{he}}$ or $\vec{\textbf{she}}$ then we label that word gender-specific. If both the words are far, then we label as gender-neutral.



\begin{equation}
  d_h = cos(\vec{\textbf{x}},\vec{\textbf{he}}) < \tau_1 \implies \vec{\textbf{x}} \equiv man  
\end{equation}
\begin{equation}
  d_w = cos(\vec{\textbf{y}},\vec{\textbf{she}}) < \tau_2 \implies \vec{\textbf{y}} \equiv woman  
\end{equation}
\begin{equation}
  d_h \geq \tau_1\ \&\ d_w \geq \tau_2 \implies \text{gender-neutral pair}  
\end{equation}


\item \textbf{Action and Relationship Extraction from Test Data} - After we have gender specific and gender neutral words from the fact data, we extract actions and relationships associated with books characters, from the test data. We extract the gender information for each characters in the books by using baby names census lists, and using this information we perform co-referencing on the textual description using \emph{Stanford OpenIE} \cite{fader2011identifying}. Next, we collate actions and relationships corresponding to each character.

\item \textbf{Bias detection using Actions} - At this point we have the actions extracted from biased data corresponding to each gender. We can now use this data against fact data to check for bias, which is shown in our demo.  

\item \textbf{Bias Removal} - To ensure making practically meaningful exchanges (e.g., exchange a {\it prominent} male character with a {\it prominent} female character for practicality), we construct a knowledge graph for each character using relations from \emph{Stanford dependency parser}. We use this graph to calculate the between-ness centrality for each character, and interchange only pairs where the centrality scores are within an empirically set threshold.

\end{asparaenum}


\begin{figure}
\centering
 \includegraphics[width=\linewidth]{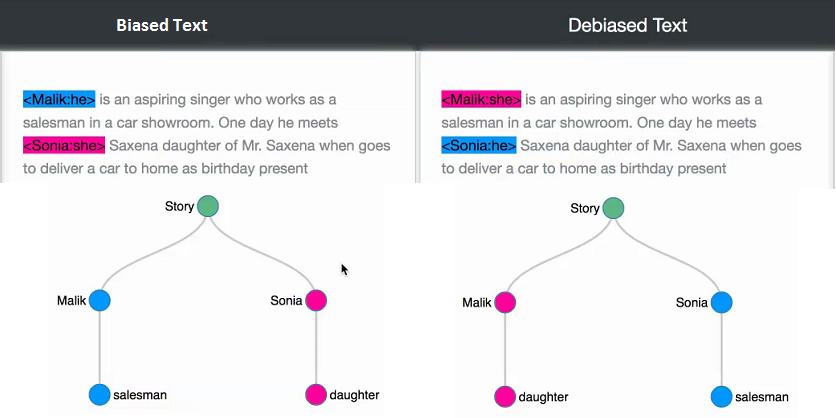}
    \caption{Text is de-biased and knowledge-graph is visualized.}
    \label{fig:screen2}
\end{figure}

\section{Conclusion}
This paper presents an analysis study which aims to extract existing gender stereotypes and biases from Man Bookers Prize Winning fiction data containing ~275 books. The analysis is performed at sentence at multi-sentence level studying the bias in data. We observed that while analyzing occupations for males and females, higher level roles are designated to males while lower level roles are designated to females. We use this rich information extracted from Goodreads to study the dynamics of the data and to further define new ways of removing such biases present in the data.
As a part of future work, we aim to extract summaries from this data which are bias-free. In this way, the next generations would stop inheriting bias from previous generations. 

\bibliography{acl2018}
\bibliographystyle{acl_natbib}
\end{document}